\documentclass[runningheads]{llncs}
\usepackage{graphicx, float, wrapfig}
\usepackage{comment}
\usepackage{amsmath,amssymb, amsfonts} 
\usepackage{color}
\usepackage{caption}
\usepackage{url}
\usepackage{enumitem, mathtools}
\usepackage{booktabs, multirow, makecell, boldline}
\usepackage[colorlinks=true]{hyperref}
\graphicspath{{figs/}}

\begin{document}

\vspace{30mm}
{ \large
\begin{itemize}[leftmargin=0.5cm, align=parleft, labelsep=2.5cm, itemsep=5ex,]

\item[\textbf{Citation}]{G. Kwon, M. Prabhushankar, D. Temel, and G. AIRegib, “Backpropagated Gradient Representations for Anomaly Detection,” In Proceedings of the European Conference on Computer Vision (ECCV), 2020.}

\item[\textbf{Review}]{Date of Publication: August 23, 2020}

\item[\textbf{Codes}]{\url{https://github.com/olivesgatech/gradcon-anomaly}}

\item[\textbf{Bib}]  {@inproceedings\{kwon2020backpropagated,\\
    title=\{Backpropagated Gradient Representations for Anomaly Detection\},\\
    author=\{Kwon, Gukyeong and Prabhushankar, Mohit and Temel, Dogancan and AlRegib, Ghassan\},\\
    booktitle=\{Proceedings of the European Conference on Computer Vision (ECCV)\},\\
    year=\{2020\}\}}

\item[\textbf{Contact}]{
\{gukyeong.kwon, mohit.p, cantemel, alregib\}@gatech.edu\\
\url{https://ghassanalregib.info/}\\}
\end{itemize}
}
\newpage
\clearpage
\setcounter{page}{1}

\pagestyle{headings}
\mainmatter
\def\ECCVSubNumber{3689}  

\title{Backpropagated Gradient Representations\\for Anomaly Detection} 

\titlerunning{Backpropagated Gradient Representations for Anomaly Detection}
%
\author{Gukyeong Kwon \and
Mohit Prabhushankar \and Dogancan Temel \and\\ Ghassan AlRegib}
\authorrunning{G. Kwon et al.}
%
\institute{Georgia Institute of Technology, Atlanta, GA 30332, USA\\
\email{\{gukyeong.kwon, mohit.p, cantemel, alregib\}@gatech.edu}}
\maketitle

\begin{abstract}
Learning representations that clearly distinguish between normal and abnormal data is key to the success of anomaly detection. Most of existing anomaly detection algorithms use activation representations from forward propagation while not exploiting gradients from backpropagation to characterize data. Gradients capture model updates required to represent data. Anomalies require more drastic model updates to fully represent them compared to normal data. Hence, we propose the utilization of backpropagated gradients as representations to characterize model behavior on anomalies and, consequently, detect such anomalies. We show that the proposed method using gradient-based representations achieves state-of-the-art anomaly detection performance in benchmark image recognition datasets. Also, we highlight the computational efficiency and the simplicity of the proposed method in comparison with other state-of-the-art methods relying on adversarial networks or autoregressive models, which require at least 27 times more model parameters than the proposed method. 

\vspace{-0.2cm}
\end{abstract}
\vspace{-0.8cm}
\section{Introduction}\label{sec:intro}
\vspace{-0.3cm}Recent advancements in deep learning enable algorithms to achieve state-of-the-art performance in diverse applications such as image classification, image segmentation, and object detection. However, the performance of such learning algorithms still suffers when abnormal data is given to the algorithms. Abnormal data encompasses data whose classes or attributes differ from training samples. Recent studies have revealed the vulnerability of deep neural networks against abnormal data~\cite{schlegl2017unsupervised},~\cite{zong2018deep}. This becomes particularly problematic when trained models are deployed in critical real-world scenarios. The neural networks can make wrong prediction for anomalies with high confidence and lead to vital consequences. Therefore, understanding and detecting abnormal data are significantly important research topics.  

\begin{figure*}[t]
    \centering
 	\includegraphics[width=0.65\linewidth]{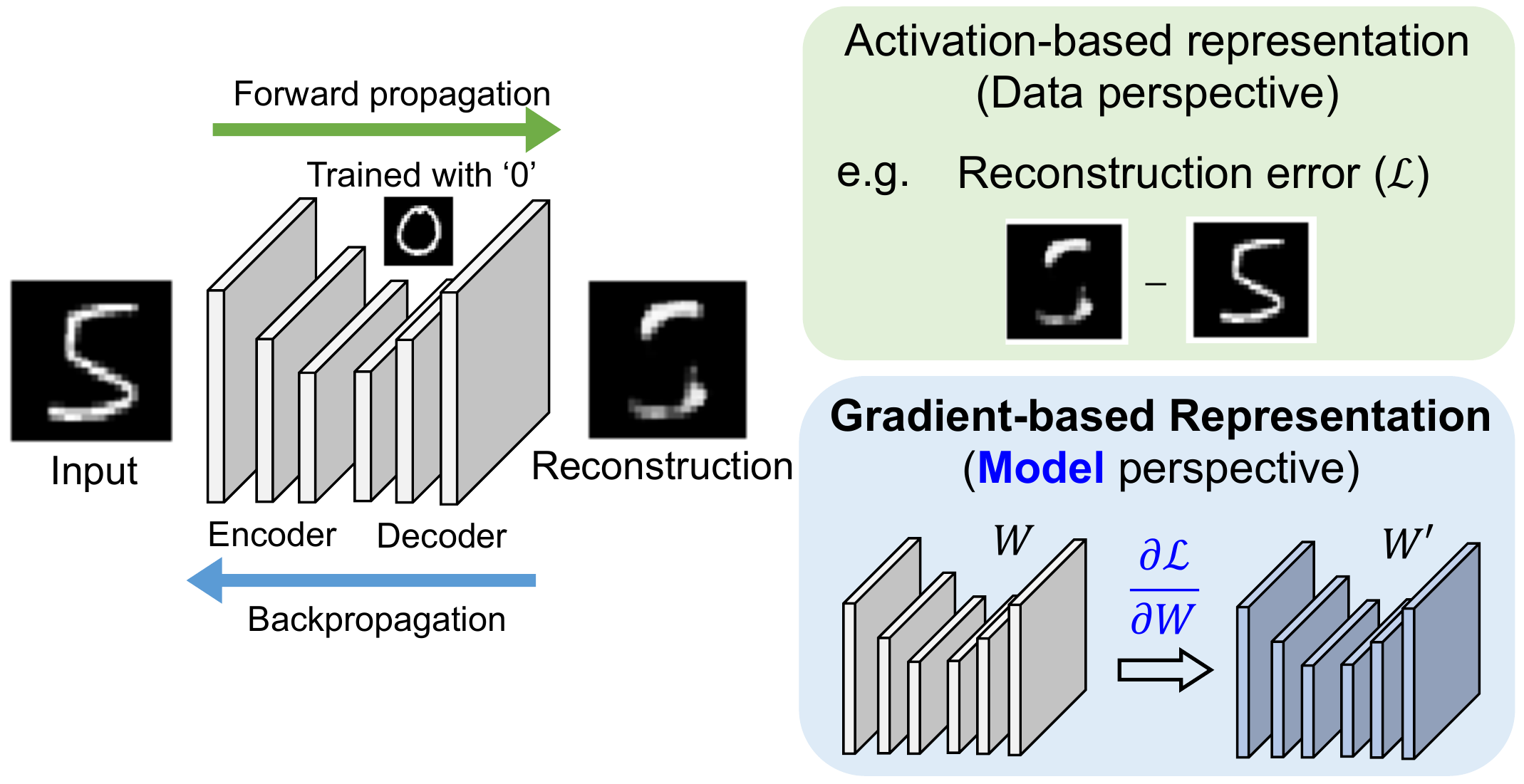}
 	\vspace{-0.3cm}
 	\caption{Activation and gradient-based representation for anomaly detection. While activation characterizes how much of input correspond to learned information, gradients focus on model updates required by the input.}\label{fig:intro}
 	\vspace{-0.7cm}
\end{figure*}

Representation from neural networks plays a key role in anomaly detection. The representation is expected to clearly differentiate normal data from abnormal data. To achieve the separation, most of existing anomaly detection algorithms deploy a representation obtained in a form of activation. The activation-based representation is constrained during training. During inference, deviation of activation from the constrained representation is formulated as an anomaly score. In Fig.~\ref{fig:intro}, we demonstrate an example of a widely used activation-based representation from an autoencoder. Assume that the autoencoder is trained with digit `0' and learns to accurately reconstruct curved edges. When an abnormal image, digit `5', is given to the network, the top and bottom curved edges are correctly reconstructed but the relatively complicated structure of straight edges in the middle cannot be reconstructed. Reconstruction error measures the difference between the target and the reconstructed image and it can be used to detect anomalies~\cite{abati2019latent},~\cite{zhou2017anomaly}. The reconstructed image, which is the activation-based representation from the autoencoder, characterizes what the network knows about input. Thus, abnormality is characterized by measuring \textit{how much of the input does not correspond to the learned information of the network}.  

In this paper, we propose using gradient-based representations to detect anomalies by characterizing model updates caused by data. Gradients are generated through backpropagation to train neural networks by minimizing designed loss functions~\cite{rumelhart1986learning}. During training, the gradients with respect to the weights provide directional information to update the neural network and learn knowledge that it has not learned. The gradients from normal data do not guide a significant change of the current weight. However, the gradients from abnormal data guide more drastic updates on the network to fully represent data. In the example given in Fig.~\ref{fig:intro}, the autoencoder needs larger updates to accurately reconstruct the abnormal image, digit `5', than the normal image, digit `0'. Therefore, the gradients can be utilized as representations to characterize abnormality of data. We propose to detect anomalies by measuring \textit{how much model update is required by the input compared to normal data}.

The gradient-based representations have several advantages compared to the activation-based representations, particularly for anomaly detection. First of all, the gradient-based representations provide abnormality characterization at different levels of data abstraction. The deviation of the activation-based representations from the constraint, often formulated as a loss ($\mathcal{L}$), is measured from the output of specific layers. On the other hand, the gradients with respect to the weights ($\frac{\partial \mathcal{L}}{\partial \mathcal{W}}$) can be obtained from any layer through backpropagation. This enables the algorithm to capture fine-grained abnormality both in low-level characteristics such as edge or color and high-level class semantics. In addition, the gradient-based representations provide directional information to characterize anomalies. The loss in the activation-based representation often measures the distance between representations of normal and abnormal data. However, by utilizing a loss defined in the gradient-based representations, we can use vectors to analyze direction in which the representation of abnormal data deviates from that of normal data. Considering that the gradients are obtained in parallel with the activation, the directional information of the gradients provides complementary features for anomaly detection along with the activation.

The gradients as representations have not been actively explored for anomaly detection. The gradients have been utilized in diverse applications such as adversarial attack generation and visualization~\cite{zeiler2014visualizing}, \cite{goodfellow2014explaining}. However, to the best of our knowledge, this paper is the first attempt to explore the representation capability of backpropagated gradients for anomalies. We provide a theoretical explanation for using gradient-based representations to detect anomalies based on the theory of information geometry, particularly using Fisher kernel principal~\cite{jaakkola1999exploiting}. In addition, through comprehensive experiments with activation-based representations, we validate the effectiveness of gradient-based representations in abnormal class and condition detection, which aims at detecting data from unseen classes and abnormal conditions. We show that the proposed anomaly detection algorithm using the gradient-based representations achieves state-of-the-art performance. The main contributions of this paper are three folds:\vspace{-0.15cm} 
\begin{enumerate}[label=\roman*, leftmargin=0.5cm]
    \item We propose utilizing backpropagated gradients as representations to characterize anomalies.
    \item We validate the representation capability of gradients for anomaly detection in comparison with activation through comprehensive baseline experiments.
    \item We propose an anomaly detection algorithm using gradient-based representations and show that it outperforms state-of-the-art algorithms using activation-based representations.
\end{enumerate}\vspace{-0.45cm}

\section{Related Works}\label{sec:related}
\vspace{-0.23cm}
\subsection{Anomaly Detection}\vspace{-0.13cm}
Most of the existing anomaly detection algorithms are focused on learning constrained activation-based representations during training. Several works propose to directly learn hyperplane or hypersphere in hidden representation space to detect anomalies. One-Class support vector machine (OC-SVM) learns a maximum margin hyperplane which separates data from the origin in the feature space~\cite{OCSVM}. Abnormal data is expected to lie on the other side of normal data and separated by the hyperplane. The authors in~\cite{tax2004support} extend the idea of OC-SVM and propose to learn a smallest hypersphere that encloses the most of training data in the feature space. In~\cite{ruff2018deep}, a deep neural network is trained to constrain the activation-based representations of data into the minimum volume of hypersphere. For a given test sample, an anomaly score is defined by the distance between the sample and the center of hypersphere.

An autoencoder has been a dominant learning framework for anomaly detection. The autoencoder generates two well-constrained representations, which are latent representation and reconstructed image representation. Based on these constrained representations, latent loss or reconstruction error have been widely used as anomaly scores. In~\cite{sakurada2014anomaly}, \cite{zhou2017anomaly}, the authors argue that anomalies cannot be accurately projected in the latent space and are poorly reconstructed. Therefore, they propose to use the reconstruction error to detect anomalies. The authors in~\cite{zong2018deep} fit Gaussian mixture models (GMM) to reconstruction error features and latent variables and estimate the likelihood of inputs to detect anomalies. In~\cite{abati2019latent}, the authors develop an autoregressive density estimation model to learn the probability distribution of the latent representation. The likelihood of the latent representation and the reconstruction error are used to detect abnormal data. 

Adversarial training is also actively explored to differentiate the representation of abnormal data. In general, a generator learns to generate realistic data similar to training data and a discriminator is trained to discriminate whether the data is generated from the generator (fake) or from training data (real)~\cite{goodfellow2014generative}. The discriminator learns a decision boundary around training data and is utilized as an abnormality detector during testing. In~\cite{sabokrou2018adversarially}, the authors adversarilally train a discriminator with an autoencoder to classify reconstructed images from original images and distorted images. The discriminator is utilized as an anomaly detector during testing. In~\cite{schlegl2017unsupervised}, the mapping from a query image to a latent variable in a generative adversarial network (GAN)~\cite{goodfellow2014generative} is estimated. The loss which measures visual similarity and feature matching for the mapping is utilized as an anomaly score. The authors in~\cite{pidhorskyi2018generative} use an adversarial autoencoder~\cite{makhzani2015adversarial} to learn the parameterized manifold in the latent space and estimate probability distributions for anomaly detection.

Aforementioned works exclusively focus on distinguishing between normal and abnormal data in the activation-based representations. In particular, most of the algorithms use adversarial networks or likelihood estimation networks to further constrain activation-based representations. These networks often require a large amount of training parameters and computations. We show that a directional constraint imposed on the gradient-based representations enables to achieve the state-of-the-art anomaly detection performance using only a backbone autoencoder with significantly less number of model parameters.\vspace{-0.2cm} 

\vspace{-0.1cm}
\subsection{Backpropagated Gradients}\vspace{-0.1cm}
The backpropagated gradients have been utilized in diverse applications including but not limited to visualization, adversarial attacks, and image classification. The backpropagated gradients have been widely used for the visualization of deep networks. In~\cite{zeiler2014visualizing}, \cite{springenberg2014striving}, information that networks have learned for a specific target class is mapped back to the pixel space through the backpropagation and visualized. The authors in~\cite{selvaraju2017grad} utilize the gradients with respect to the activation to weight the activation and visualize the reasoning for prediction that neural networks have made. An adversarial attack is another application of gradients. In~\cite{goodfellow2014explaining}, \cite{kurakin2016adversarial}, the authors show that adversarial attacks can be generated by adding an imperceptibly small vector which is the signum of input gradients. Several works have incorporated gradients with respect to the input in a form of regularization during the training of neural networks to improve the robustness~\cite{drucker1991double}, \cite{ross2018improving}, \cite{sokolic2017robust}. Although existing works have shown that the gradients with respect to the input or the activation can be useful for diverse applications, the gradients with respect to the weights of neural networks have not been actively explored aside from its role in training deep networks. 

A few works have explored the gradients with respect to the model parameters as features for data. The authors in~\cite{perronnin2007fisher} propose to use Fisher kernels which are based on the normalized gradient vectors of the generative model for image categorization. The authors in~\cite{achille2019information}, \cite{achille2019task2vec} characterize information encoded in the neural network and utilize Fisher information to represent tasks. In~\cite{kwon2019distorted}, the gradients of the neural network are utilized to classify distorted images and objectively estimate the quality of them.
The gradients have been also studied as a local liner approximation to a neural network~\cite{mu2020gradients}. Our approach differs from other existing works in two main aspects. First, we generalize the Fisher kernel principal using the backpropagated gradients from the neural networks. Since we use the backpropagated gradients to estimate the Fisher score of normal data distribution, the data does not need to be modeled by known probabilistic distributions such as a GMM. Second, we use the gradients to represent information that the networks have not learned. In particular, we provide our interpretation of gradients which characterize abnormal information for the neural networks and validate their effectiveness in anomaly detection.\vspace{-0.3cm}

\section{Gradient-based Representations}\label{sec:geometric}
\vspace{-0.2cm}In this section, the intuition to using gradient-based representation for anomaly detection is detailed. In particular, we present our interpretation of gradients from a geometric and a theoretical perspective. Geometric interpretation of gradients highlights the advantages of the gradients over activation from a data manifold perspective. Also, theory of information geometry further supports the characterization of anomalies using the gradients.\vspace{-0.3cm}

\subsection{Geometric Interpretation of Gradients}
\label{subsec:dataManifold}\vspace{-0.2cm}
We use an autoencoder, which is an unsupervised representation learning framework to explain the geometric interpretation of gradients. An autoencoder consists of an encoder, $f_\theta$, and a decoder, $g_\phi$. From an input image, $x$, a latent variable, $z$, is generated as $z = f_\theta (x)$ and a reconstructed image is obtained by feeding the latent variable into the decoder, $g_\phi(f_\theta(x))$. The training is performed by minimizing a loss function, $J(x; \theta, \phi)$, defined as follows:\vspace{-0.2cm}
\begin{equation}\label{eq:loss}
    J(x; \theta, \phi) = \mathcal{L}(x, g_{\phi}(f_{\theta}(x))) + \Omega(z; \theta, \phi),\vspace{-0.2cm}
\end{equation}
where $\mathcal{L}$ is a reconstruction error, which measures the dissimilarity between the input and the reconstructed image and $\Omega$ is a regularization term for the latent variable.

\begin{figure}[t]
\begin{minipage}{0.67\linewidth}
    \vspace{0cm}
    \centering
 	\includegraphics[width=\linewidth]{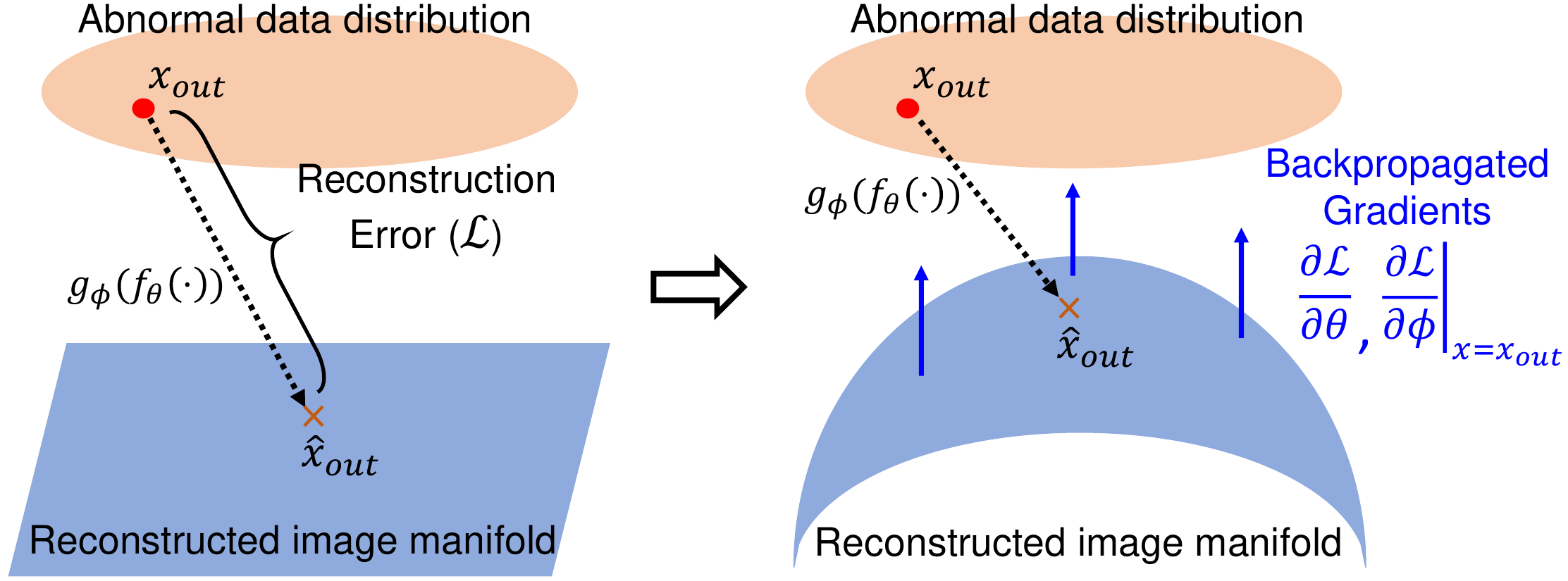}\vspace{-0.3cm}
 	\caption{Geometric interpretation of gradients.}\label{fig:manifold}
\end{minipage}
\begin{minipage}{0.32\linewidth}
    \centering
	\includegraphics[width=\linewidth]{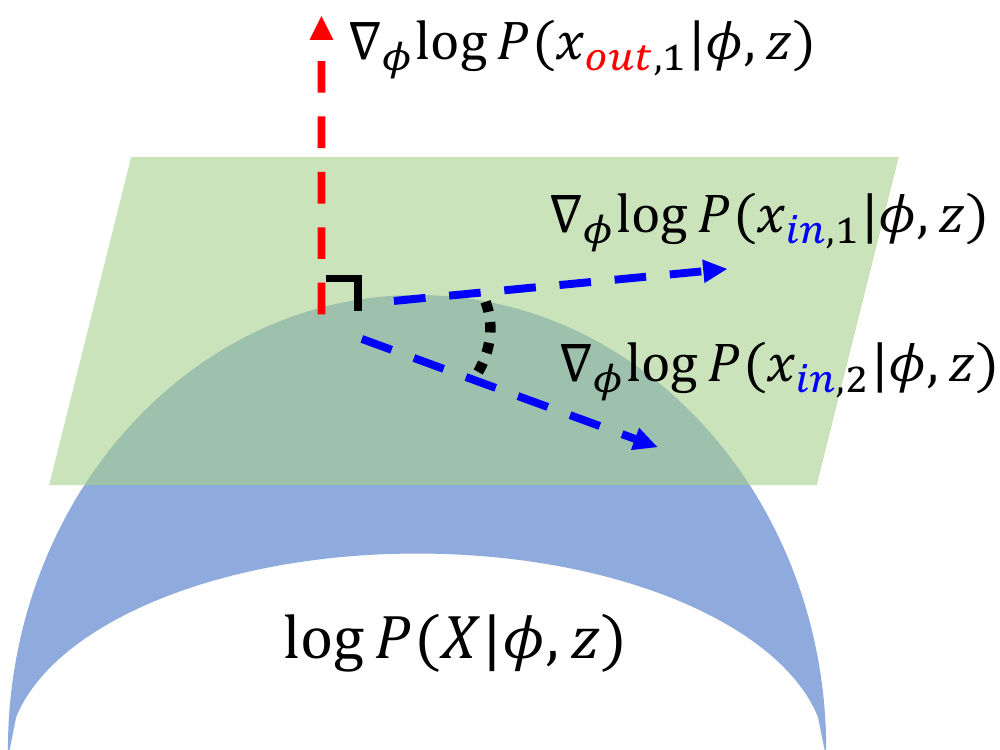}\vspace{-0.3cm}
	\caption{Gradient constraint on the manifold.}\label{fig:constraint}
\end{minipage}\vspace{-0.7cm}
\end{figure}
We visualize the geometric interpretation of backpropagated gradients in Fig.~\ref{fig:manifold}. The autoencoder is trained to accurately reconstruct training images and the reconstructed training images form a manifold. We assume that the structure of the manifold is a linear plane as shown in the figure for the simplicity of explanation. During testing, any given input to the autoencoder is projected onto the reconstructed image manifold through the projection, $g_\phi(f_\theta(\cdot))$. Ideally, perfect reconstruction is achieved when the reconstructed image manifold includes the input image. Assume that abnormal data distribution is outside of the reconstructed image manifold. When an abnormal image, $x_{out}$, sampled from the distribution is input to the autoencoder, it will be reconstructed as $\hat{x}_{out}$ through the projection, $g_\phi(f_\theta(x_{out}))$. Since the abnormal image has not been utilized for training, it will be poorly reconstructed. The distance between $x_{out}$ and $\hat{x}_{out}$ is formulated as the reconstruction error and characterizes the abnormality of the data as shown in the left side of Fig.~\ref{fig:manifold}. The gradients with respect to the weights, $\frac{\partial \mathcal{L}}{\partial \theta}, \frac{\partial \mathcal{L}}{\partial \phi}$, can be calculated through the backpropagation of the reconstruction error. These gradients represent required changes in the reconstructed image manifold to incorporate the abnormal image and reconstruct it accurately as shown in the right side of Fig.~\ref{fig:manifold}. In other words, these gradients characterize orthogonal variations of the abnormal data distribution with respect to the reconstructed image manifold.

The interpretation of gradients from the data manifold perspective highlights the advantages of gradients in anomaly detection. In activation-based representations, the abnormality is characterized by distance information measured using a designed loss function. On the other hand, the gradients provide directional information, which indicates the movement of manifold in which data representations reside. This movement characterizes, in particular, in which direction the abnormal data distribution deviates from the representations of normal data. Furthermore, the gradients obtained from different layers provide a comprehensive perspective to represent anomalies with respect to the current representations of normal data. Therefore, the directional information from gradients can be utilized as complementary information to the distance information from the activation.

\vspace{-0.4cm}
\subsection{Theoretical Interpretation of Gradients}\label{subsec:Fisher}\vspace{-0.2cm}
We derive theoretical explanation for gradient-based representations from information geometry, particularly using the Fisher kernel. Based on the Fisher kernel, we show that the gradient-based representations characterize model updates from query data and differentiate normal from abnormal data. We utilize the same setup of an autoencoder described in Section~\ref{subsec:dataManifold} but consider the encoder and the decoder as probability distributions~\cite{goodfellow2016deep}. Given the latent variable, $z$, the decoder models input distribution through a conditional distribution, $P_\phi(x|z)$. The autoencoder is trained to minimize the negative log-likelihood, $-\log P_\phi (x|z)$. When $x$ is a real value and $P_\phi(x|z)$ is assumed to be a Gaussian distribution, the decoder estimates the mean of the Gaussian. Also, the minimization of the negative log-likelihood corresponds to using a mean squared error as the reconstruction error. When $x$ is a binary value, the decoder is assumed to be a Bernoulli distribution. The negative log-likelihood is formulated as a binary cross entropy loss. Considering the decoder as the conditional probability enables to interpret gradients using the Fisher kernel.

The Fisher kernel defines a metric between samples using the gradients of generative probability distribution~\cite{jaakkola1999exploiting}. Let $X$ be a set of samples and $P(X|\theta)$ is a probability density function of the samples parameterized by $\theta=[\theta_1, \theta_2, ..., \theta_N]^\mathsf{T} \in \mathbb{R}^N$. This probability distribution models a Riemannian manifold with a local metric defined by Fisher information matrix, $F \in \mathbb{R}^{N \times N}$, as follows:\vspace{-0.2cm}
\begin{equation}
    F = \mathop{\mathbb{E}}_{x \in X}[U_{\theta}^X {U_{\theta}^X}^\mathsf{T}] \quad \text{where} \quad U_{\theta}^X = \nabla_\theta \log P(X|\theta).\vspace{-0.3cm}
\end{equation}
$U_{\theta}^X$ is called the Fisher score which describes the contribution of the parameters in modeling the data distribution. In~\cite{jaakkola1999exploiting}, the authors propose the Fisher kernel to measure the difference between two samples based on the Fisher score. The Fisher kernel, $K_{FK}$, is defined as\vspace{-0.3cm}
\begin{equation}
    K_{FK} (X_i, X_j) = {{U_{\theta}}^{X_i}}^\mathsf{T} F^{-1} U_{\theta}^{X_{j}},\vspace{-0.2cm}
\end{equation}
where $X_i$ and $X_j$ are two data samples. The Fisher kernels enable to extract discriminant features from the generative model and they have been actively used in diverse applications such as image categorization, image classification, and action recognition~\cite{perronnin2007fisher}, \cite{sanchez2013image}, \cite{peng2014action}.

We use the Fisher kernel estimated from the autoencoder for anomaly detection. The distribution of the decoder is parameterized by the weights, $\phi$, and the Fisher score from the decoder is defined as $U_{\phi, z}^X = \nabla_\phi \log P(X|\phi, z)$. Also, since the distribution is learned to be generalizable to the test data, we can use the Fisher kernel to measure the distance between training data and normal test data, and between training data and abnormal test data. The Fisher kernel for normal data (inliers), $K_{FK}^{in}$, and abnormal data (outliers), $K_{FK}^{out}$, are derived as follows, respectively:\vspace{-0.4cm}
\begin{equation}
    K_{FK}^{in} (X_{tr}, X_{te,in}) = {{U_{\phi}}^{X_{tr}}}^\mathsf{T} F^{-1} U_{\phi, z}^{X_{te,in}}
\end{equation}\vspace{-0.5cm}
\begin{equation}
    K_{FK}^{out} (X_{tr}, X_{te,out}) = {{U_{\phi}}^{X_{tr}}}^\mathsf{T} F^{-1} U_{\phi, z}^{X_{te,out}},\vspace{-0.1cm}
\end{equation}
where $X_{tr}, X_{te, in}, X_{te, out}$ are training data, normal test data, and abnormal test data, respectively. For ideal anomaly detection, $K_{FK}^{out}$ should be larger than $K_{FK}^{in}$ to clearly differentiate normal and abnormal
data. The difference between $K_{FK}^{in}$ and $K_{FK}^{out}$ is characterized by the Fisher scores $U_{\phi, z}^{X_{te,in}}$ and $U_{\phi, z}^{X_{te,out}}$. Therefore, the Fisher scores from query data are discriminant features for detecting anomalies. We propose to estimate the Fisher scores using the backpropagated gradients with respect to the weights of the decoder. Since the autoencoder is trained to minimize the negative log-likelihood loss, $\mathcal{L} = -\log P_\phi (x|z)$, the backpropagated gradients, $\frac{\partial \mathcal{L}}{\partial \phi}$, obtained from normal and abnormal data estimate $U_{\phi, z}^{X_{te,in}}$ and $U_{\phi, z}^{X_{te,out}}$ when the autoencoder is trained with a sufficiently large amount of data to model the data distribution. Therefore, we can interpret the gradient-based representations as discriminant representations obtained from the conditional probabilistic modeling of data for anomaly detection.

We visualize the gradients with respect to the weights of the decoder obtained by backpropagating the reconstruction error, $\mathcal{L}$, from normal data, $x_{in,1}, x_{in,2}$, and abnormal data, $x_{out, 1}$, in Fig.~\ref{fig:constraint}. These gradients estimate the Fisher scores for inliers and outliers, which need to be clearly separated for anomaly detection. Given the definition of the Fisher scores, the gradients from normal data should contribute less to the change of the manifold compared to those from abnormal data. Therefore, the gradients from normal data should reside in the tangent space of the manifold but abnormal data results in the gradients orthogonal to the tangent space. We achieve this separation in gradient-based representations through directional constraint described in the following section.\vspace{-0.3cm}

\section{Method: Gradient Constraint}\label{sec:const}
\vspace{-0.1cm}The separation between inliers and outliers in the representation space is often achieved by modeling the normality of data. The deviation from the normality model captures the abnormality. The normality is often modeled through constraints imposed during training. The constraint allows normal data to be easily constrained but makes abnormal data deviates. For example, the autoencoders constrain the output to be similar to the input and the reconstruction error measures the deviation. A variational autoencoder (VAE)~\cite{VAE} and an adversarial autoencoder (AAE) often constrain the latent representation to follow the Gaussian distribution and the deviation from the Gaussian distribution characterizes anomalies. In the gradient-based representations, we also impose a constraint during training to model the normality of data and further differentiate $U_{\phi, z}^{X_{te,in}}$ from $U_{\phi, z}^{X_{te,out}}$ defined in Section~\ref{subsec:Fisher}.

We propose to train an autoencoder with a directional gradient constraint to model the normality. In particular, based on the interpretation of gradients from the Fisher kernel perspective, we enforce the alignment between gradients. This constraint makes the gradients from normal data aligned with each other and result in small changes to the manifold. On the other hand, the gradients from abnormal data will not be aligned with others and guide abrupt changes to the manifold. We utilize a gradient loss, $\mathcal{L}_{grad}$, as a regularization term in the entire loss function, $J$. We calculate the cosine similarity between the gradients of a certain layer $i$ in the decoder at the $k^{th}$ iteration of training, $\frac{\partial \mathcal{L}}{\partial \phi_i}^{k}$, and the average of the training gradients of the same layer $i$ obtained until the $(k-1)^{th}$ iteration, $\frac{\partial \mathcal{J}}{\partial \phi_{i}}_{avg}^{k-1}$. The gradient loss at the $k^{th}$ iteration of training is obtained by averaging the cosine similarity over all the layers in the decoder as follows:\vspace{-0.2cm}
\begin{equation}
    \mathcal{L}_{grad} = -\mathop{\mathbb{E}}_{i}\left[\text{cosSIM}\left( \dfrac{\partial \mathcal{J}}{\partial \phi_{i}}_{avg}^{k-1}, \dfrac{\partial \mathcal{L}}{\partial \phi_{i}}^{k}\right)\right], \quad  \dfrac{\partial \mathcal{J}}{\partial \phi_{i}}_{avg}^{k-1} = \dfrac{1}{\left(k -1\right)}\sum_{t = 1}^{k-1}\dfrac{\partial \mathcal{J}}{\partial \phi_{i}}^{t},\label{eq:cosSIM}
\end{equation}
where $J$ is defined as $J = \mathcal{L} + \Omega + \alpha\mathcal{L}_{grad}$. The first and the second terms are the reconstruction error and the latent loss, respectively and they are defined by different types of autoencoders. $\alpha$ is a weight for the gradient loss. We set sufficiently small $\alpha$ value to ensure that gradients actively explore the optimal weights until the reconstruction error and the latent loss become small enough. Based on the interpretation of the gradients described in Section~\ref{subsec:Fisher}, we only constrain the gradients of the decoder layers and the encoder layers remain unconstrained. 

During training, $\mathcal{L}$ is first calculated from the forward propagation. Through the backpropagation, $\frac{\partial \mathcal{L}}{\partial \phi_{i}}^{k}$ is obtained without updating the weights. Based on the obtained gradient, the entire loss $J$ is calculated and finally the weights are updated using backpropagated gradients from the loss $J$. An anomaly score is defined by the combination of the reconstruction error and the gradient loss as $\mathcal{L} + \beta \mathcal{L}_{grad}$. Although we use $\alpha$ to weight the gradient loss during training, we found that the gradient loss is often more effective than the reconstruction error for anomaly detection. To better balance the two losses, we use $\beta = 4 \alpha$ for all the experiments and show that the weighted combination of two losses improve the performance. The proposed anomaly detection algorithm using \textbf{Grad}ient \textbf{Con}straint is called GradCon.\vspace{-0.2cm}

\section{Experiments}\label{sec:exp}
\vspace{-0.15cm}\subsection{Experimental Setup}\label{subsec:setup}\vspace{-0.15cm}
We conduct anomaly detection experiments to both qualitatively and quantitatively evaluate the performance of the gradient-based representations. In particular, we perform abnormal class detection and abnormal condition detection using the gradient constraint and compare \texttt{GradCon} with other state-of-the-art activation-based anomaly detection algorithms. In abnormal class detection, images from one class of a dataset are considered as inliers and used for the training. Images from other classes are considered as outliers. In abnormal condition detection, images without any effect are utilized as inliers and images captured under challenging conditions such as distortions or environmental effects are considered as outliers. Both inliers and outliers are given to the network during testing. The anomaly detection algorithms are expected to correctly classify data of which class and condition differ from those of the training data.

\noindent\textbf{Datasets} We utilize four benchmark datasets, which are CIFAR-10~\cite{krizhevsky2009learning}, MNIST~\cite{lecun1998gradient}, fashion MNIST (fMNIST)~\cite{xiao2017fashion}, and CURE-TSR~\cite{temel2017cure} to evaluate the performance of the proposed algorithm. We use CIFAR-10, MNIST, fMNIST for abnormal class detection and CURE-TSR for abnormal condition detection. CIFAR-10 dataset consists of 60,000 color images with 10 classes. MNIST dataset contains 70,000 handwritten digit images from 0 to 9 and fMNIST dataset also has 10 classes of fashion products and there are 7,000 images per class. CURE-TSR dataset has $637,560$ color traffic sign images which consist of 14 traffic sign types under 5 levels of 12 different challenging conditions. For CIFAR-10, CURE-TSR, and MNIST, we follow the protocol described in~\cite{perera2019ocgan} to create splits. To be specific, we utilize the original training and the test split of each dataset for training and testing. $10\%$ of training images are held out for validation. For fMNIST, we follow the protocol described in~\cite{pidhorskyi2018generative}. The dataset is split into 5 folds and  $60\%$ of each class is used for training, 20\% is used for validation, the remaining 20\% is used for testing. In the experiments with CIFAR-10, MNIST, and fMNIST, we use images from one class as inliers for training. During testing, inlier images and the same number of oulier images randomly sampled from other classes are utilized. For CURE-TSR, challenge-free images are utilized as inliers for training. During testing, challenge-free images are utilized as inliers and the same images with challenging conditions are utilized as outliers. We particularly use 5 challenge levels with 8 challenging conditions which are \texttt{Decolorization}, \texttt{Lens blur}, \texttt{Dirty lens}, \texttt{Exposure}, \texttt{Gaussian blur}, \texttt{Rain}, \texttt{Snow}, and \texttt{Haze}. All the results are obtained using area under receiver operation characteristic curve (AUROC) and we also report F1 score in fMNIST dataset for the fair comparison with the state-of-the-art method~\cite{pidhorskyi2018generative}. 

\noindent\textbf{Implementation details} We train a convolutional autoencoder (CAE) for \texttt{GradCon}. The encoder and the decoder consist of 4 convolutional layers and the dimension of the latent variable is $3 \times 3 \times 64$. The number of convolutional filters for each layer in the encoder is 32, 32, 64, 64 and the kernel size is $4 \times 4$ for all the layers. The architecture of the decoder is symmetric to the encoder. Adam optimizer~\cite{kingma2014adam} with the learning rate of $0.001$ is used for training. We use mean square error as the reconstruction error and do not use any latent loss for the CAE ($\Omega = 0$). $\alpha = 0.03$ is used to weight the gradient loss.\vspace{-0.3cm}

\subsection{Baseline Comparison}\vspace{-0.15cm}
We compare the performance of the gradient-based representations in characterizing abnormal data with the activation-based representations. Furthermore, we show that the gradient-based representations can complement the activation-based representations and improve the performance of anomaly detection. We train four different autoencoders, which are CAE, CAE with the gradient constraint (CAE + Grad), VAE, VAE with the gradient constraint (VAE + Grad) for the baseline experiments. VAEs are trained using binary cross entropy as the reconstruction error and Kullback Leibler (KL) divergence as the latent loss. Implementation details for VAEs are same as those for the CAE described in Section~\ref{subsec:setup}. We train the autoencoders using images from each class of CIFAR-10. Two losses defined by the activation-based representations, which are the reconstruction error (Recon) and the latent loss (Latent), and the gradient loss (Grad) defined by the gradient-based representations are separately used as anomaly scores for detection. AUROC results are reported in Table~\ref{tab:grad_cifar} and the highest AUROC for each class is highlighted in bold. 

\noindent\textbf{Effectiveness of the gradient constraint (CAE vs. CAE+Grad)} We first compare the performance of CAE and CAE + Grad to analyze the effectiveness of the gradient-based representation with constraint. The reconstruction error from CAE and CAE + Grad achieves comparable average AUROC scores. The gradient loss from CAE + Grad achieves the best performance with an average AUROC of $0.661$. This shows that the gradient constraint marginally sacrifices the performance from the activation-based representation and achieve the superior performance from the gradient-based representation.

\begin{table}[t]
\scriptsize
\centering
\begin{tabular}{ccccccccccccc}
\toprule
Model & Loss & Plane & Car & Bird & Cat & Deer & Dog & Frog & Horse & Ship & Truck & Average \\ \hline
CAE & Recon & 0.682 & 0.353 & 0.638 & 0.587 & 0.669 & \textbf{0.613} & 0.495 & 0.498 & 0.711 & 0.390 & 0.564 \\ \hline
\multirow{2}{*}{\makecell{CAE\\ + Grad}} & Recon & 0.659 & 0.356 & \textbf{0.640} & 0.555 & 0.695 & 0.554 & 0.549 & 0.478 & 0.695 & 0.357 & 0.554 \\ \cline{2-13} 
 & Grad & \textbf{0.752} & 0.619 & 0.622 & 0.580 & 0.705 & 0.591 & 0.683 & \textbf{0.576} & \textbf{0.774} & \textbf{0.709} &\textbf{ 0.661} \\
 \hline\hline 
\multirow{2}{*}{VAE} & Recon & 0.553 & 0.608 & 0.437 & 0.546 & 0.393 & 0.531 & 0.489 & 0.515 & 0.552 & 0.631 & 0.526 \\ \cline{2-13} 
 & Latent & 0.634 & 0.442 & \textbf{0.640} & 0.497 & \textbf{0.743} & 0.515 & \textbf{0.745} & 0.527 & 0.674 & 0.416 & 0.583 \\ \hline
\multirow{3}{*}{\makecell{VAE \\ + Grad}} & Recon & 0.556 & 0.606 & 0.438 & 0.548 & 0.392 & 0.543 & 0.496 & 0.518 & 0.552 & 0.631 & 0.528 \\ \cline{2-13} 
 & Latent & 0.586 & 0.396 & 0.618 & 0.476 & 0.719 & 0.474 & 0.698 & 0.537 & 0.586 & 0.413 & 0.550 \\ \cline{2-13} 
 & Grad & 0.736 & \textbf{0.625} & 0.591 & \textbf{0.596} & 0.707 & 0.570 & 0.740 & 0.543 & 0.738 & 0.629 & 0.647 \\ \bottomrule
\end{tabular}\caption{Baseline anomaly detection results on CIFAR-10. The reconstruction error (Recon) and the latent loss (Latent) are obtained from the activation-based representations and the gradient loss (Grad) is obtained from the gradient-based representations.}\label{tab:grad_cifar}\vspace{-0.8cm}
\end{table}
\begin{figure}[t]
\begin{minipage}[t]{0.28\linewidth}
\vspace{.5cm}
  \centering
    \includegraphics[width=\linewidth]{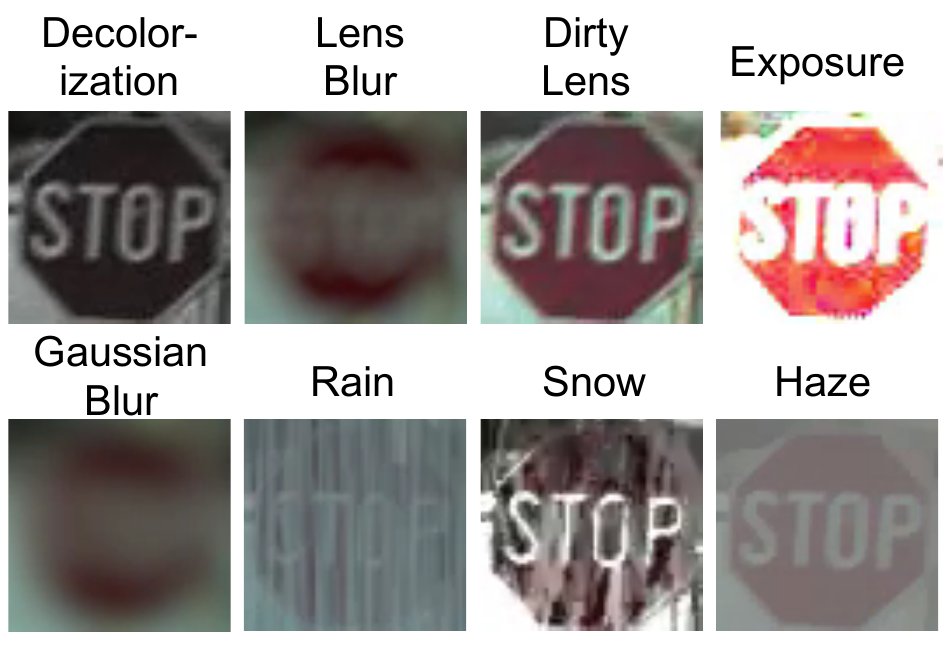}\vspace{-0.2cm}
\end{minipage}
\hspace{0.05cm}
\begin{minipage}[t]{0.7\linewidth}
\vspace{0cm}
  \centering
    \includegraphics[width=\linewidth]{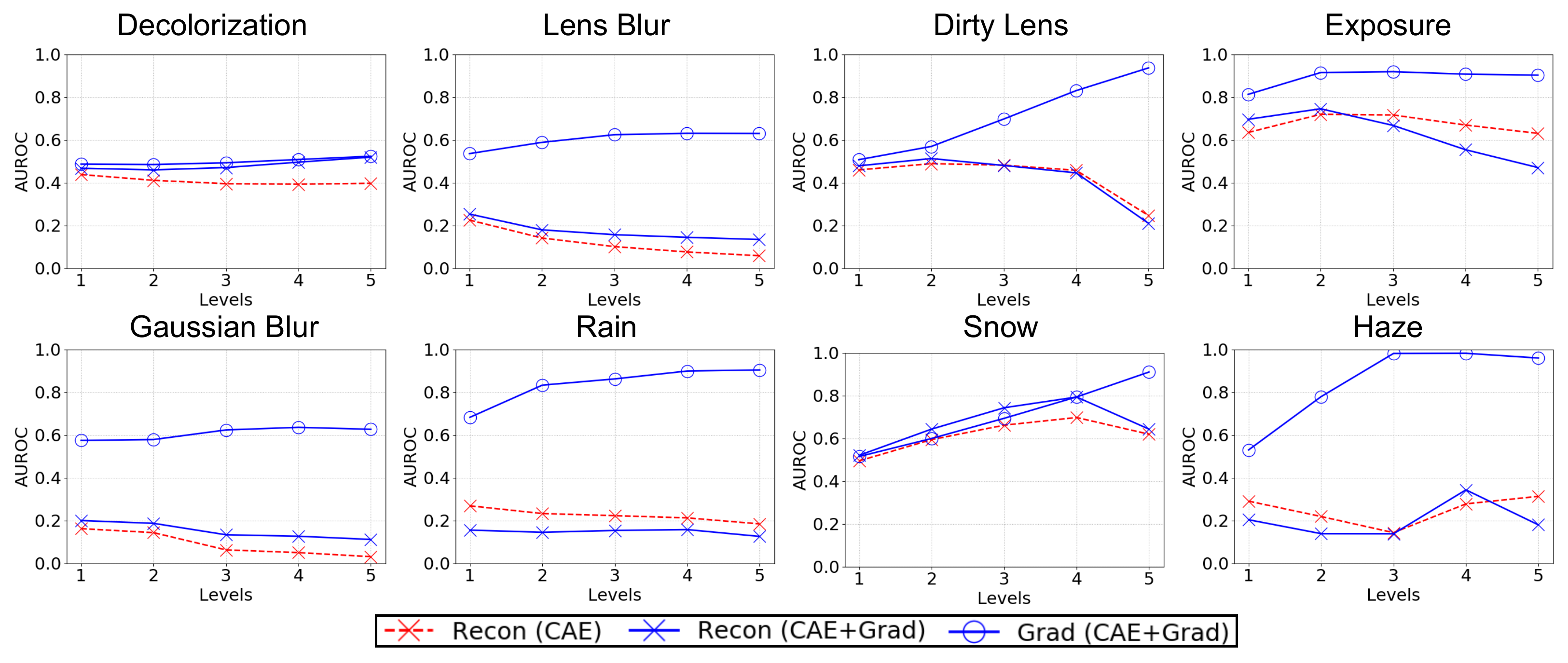}\vspace{-0.2cm}
\end{minipage}
\caption{Baseline anomaly detection results on CURE-TSR.}\label{fig:cure_results}\vspace{-0.7cm}
\end{figure}
\noindent\textbf{Performance sacrifice from the latent constraint (CAE vs. VAE)} We evaluate the effect of the latent constraint by comparing CAE and VAE. The latent loss of VAE achieves the improved performance compared to the reconstruction error of CAE by an average AUROC of $0.019$. However, the performance of the reconstruction error from VAE is lower than that from CAE by $0.038$. This shows that the latent constraint sacrifices the performance from another activation-based representation which is the reconstructed image. Since both latent representation and reconstructed image are obtained from forward propagation, the constraint imposed in the latent space affects the reconstruction performance. Therefore, using a combination of multiple activation-based representations faces limitations in improving the performance.

\noindent\textbf{Complementary features from the gradient constraint (VAE vs. VAE +Grad)} Comparison between VAE and VAE + Grad shows the effectiveness of using the gradient constraint with the activation constraint. The gradient loss in VAE + Grad achieves the second best average AUROC and outperforms the latent loss in the VAE by $0.064$. The performance from the reconstruction error is comparable between VAE and VAE + Grad. The average AUROC of the latent loss from VAE + Grad is marginally sacrificed by $0.033$ compared to that from VAE. In both CAE + Grad and VAE + Grad, the performance gain from the gradient loss is always greater than the sacrifice in other activation-based representations. This is contrary to the CAE and VAE comparison where the performance gain is smaller than the sacrifice from the reconstruction error. Since gradients are obtained in parallel with the activation, constraining gradients less affects the anomaly detection performance from the activation-based representations. Thus, the gradient-based representations can provide complementary features to the activation-based representations for anomaly detection.

\begin{table}[t]
\centering
\begin{minipage}{0.433\linewidth}
    \scriptsize
    \centering
    \begin{tabular}{ccccccc}
    \toprule
    \multicolumn{2}{c}{Layer} & $1^{st}$ & $2^{nd}$ & $3^{rd}$ & $4^{th}$ & All \\ \hline
    \multicolumn{2}{c}{CIFAR-10} & \textbf{0.648} & \textbf{0.649} & 0.628 & 0.605 & \textbf{0.661} \\ \hline
    \multirow{3}{*}{\makecell{CURE\\-TSR}} & DL & \textbf{0.688} & 0.640 & 0.649 & \textbf{0.681} & \textbf{0.708} \\ \cline{2-7} 
     & EX & \textbf{0.859} & 0.811 & 0.781 & \textbf{0.833} & \textbf{0.891} \\ \cline{2-7} 
     & SN & \textbf{0.677} & 0.612 & 0.628 & \textbf{0.693} & \textbf{0.702} \\ \bottomrule
    \end{tabular}\caption{Anomaly detection results from the gradients of each layer in the decoder.}\label{tab:layer}
\end{minipage}\hspace{0.15cm}
\begin{minipage}{0.546\linewidth}
    \centering
    \vspace{-0.1cm}
    \includegraphics[width=\linewidth]{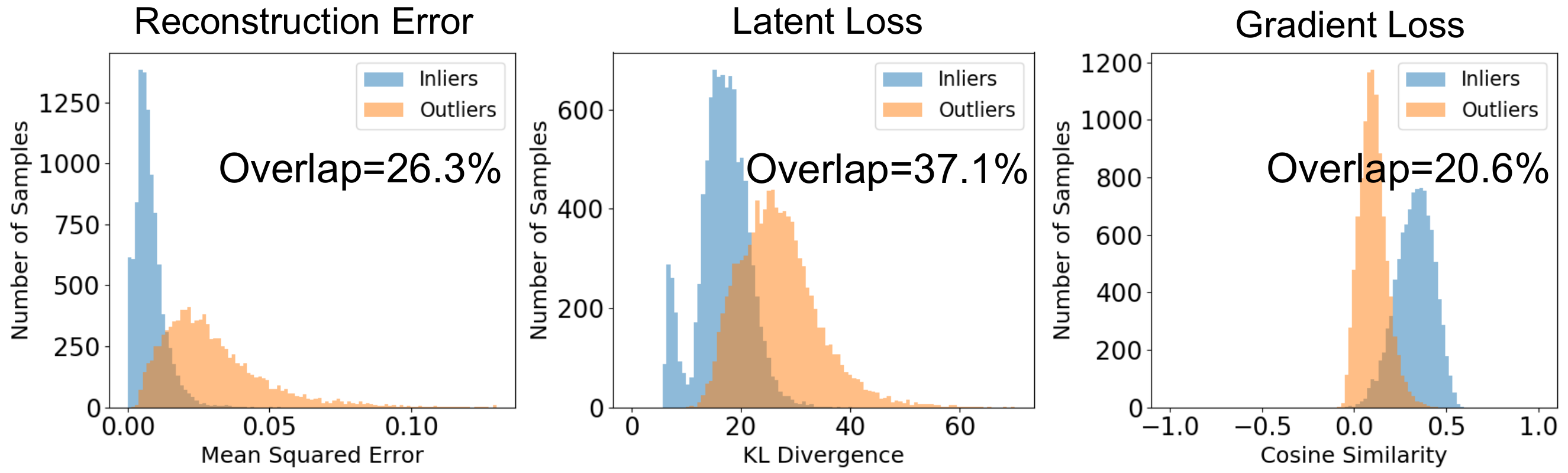}\caption*{Figure 5. Histogram analysis on activation losses and gradient loss in MNIST.}
\end{minipage}\vspace{-1cm}
\end{table}

\noindent\textbf{Abnormal condition detection} We further analyze the discriminant capability of the gradient-based representations for diverse challenging conditions and levels. We compare the performance of CAE and CAE + Grad using the reconstruction error (Recon) and the gradient loss (Grad). Samples with challenging conditions and the AUROC performance are visualized in Fig.~\ref{fig:cure_results}. For all challenging conditions and levels, CAE + Grad achieves the best performance. In particular, except for snow level 1$\sim$3, the gradient loss achieves the best performance and for snow level 1$\sim$3, the reconstruction error of CAE + Grad achieves the best performance. In terms of the average AUROC over challenge levels, the gradient loss of CAE + Grad outperforms the reconstruction error of CAE by the largest margin of $0.612$ in rain and the smallest margin of $0.089$ in snow. These test conditions encompass acquisition imperfection, processing artifact, and environmental challenging conditions. The superior performance of the gradient loss shows that the gradient-based representation effectively characterizes diverse types and levels of unseen challenging conditions.

\noindent\textbf{Decomposition of the gradient loss} We decompose the gradient loss and analyze the contribution of gradients from each layer on anomaly detection. Instead of the gradient loss obtained by averaging the cosine similarity over all the layers as~(\ref{eq:cosSIM}), we use the cosine similarity from each layer as an anomaly score. The average AUROC results obtained by the gradients from the first to the fourth layer of the decoder are reported in Table~\ref{tab:layer}. Also, results obtained by averaging the cosine similarity over all layers are reported. We use CIFAR-10 and \texttt{Dirty Lens} (DL), \texttt{Exposure} (EX), \texttt{Snow} (SN) challenge types of CURE-TSR. In CIFAR-10, inlier class and outlier classes share most of low-level features such as edges or colors. Also, semantic information mostly differentiate classes. Since the layers close to the latent space focus more on high-level characteristics of data, the gradient loss from the first and the second layer show the largest contribution on anomaly detection. In CURE-TSR, challenging conditions alter low-level characteristics of images such as edges or colors. Therefore, the last layer of the decoder also contributes more than middle layers for abnormal condition detection. This shows that gradients extracted from different layers characterize abnormality at different levels of data abstraction. In both datasets, results obtained by combining all the layers (All) show the best performance. Given that losses defined by activation-based representations can be calculated only from the output of specific layers, using gradients from all the layers enable to capture abnormality in both low-level and high-level characteristics of data.\vspace{-0.3cm}

\begin{table*}[t]
\scriptsize
\centering
\begin{tabular}{cccccccccccc}
\toprule
 & Plane & Car & Bird & Cat & Deer & Dog & Frog & Horse & Ship & Truck & Average \\ \hline
OCSVM~\cite{OCSVM} & 0.630 & 0.440 & 0.649 & 0.487 & 0.735 & 0.500 & 0.725 & 0.533 & 0.649 & 0.508 & 0.586 \\ \hline
KDE~\cite{KDE} & 0.658 & 0.520 & \textbf{0.657} & 0.497 & 0.727 & 0.496 & \textbf{0.758} & 0.564 & 0.680 & 0.540 & 0.610 \\ \hline
DAE~\cite{DAE} & 0.411 & 0.478 & 0.616 & 0.562 & 0.728 & 0.513 & 0.688 & 0.497 & 0.487 & 0.378 & 0.536 \\ \hline
VAE~\cite{VAE} & 0.634 & 0.442 & 0.640 & 0.497 & \textbf{0.743} & 0.515 & 0.745 & 0.527 & 0.674 & 0.416 & 0.583 \\ \hline
PixelCNN~\cite{PixelCNN} & \textbf{0.788} & 0.428 & 0.617 & 0.574 & 0.511 & 0.571 & 0.422 & 0.454 & 0.715 & 0.426 & 0.551 \\ \hline
LSA~\cite{abati2019latent} & 0.735 & 0.580 & \textbf{0.690} & 0.542 & \textbf{0.761} & 0.546 & \textbf{0.751} & 0.535 & 0.717 & 0.548 & 0.641 \\ \hline
AnoGAN~\cite{GAN} & 0.671 & 0.547 & 0.529 & 0.545 & 0.651 & 0.603 & 0.585 & \textbf{0.625} & 0.758 & 0.665 & 0.618 \\ \hline
DSVDD~\cite{DSVDD} & 0.617 & \textbf{0.659} & 0.508 & \textbf{0.591} & 0.609 & \textbf{0.657} & 0.677 & \textbf{0.673} & 0.759 & \textbf{0.731} & 0.648 \\ \hline
OCGAN~\cite{perera2019ocgan} & 0.757 & 0.531 & 0.640 & \textbf{0.620} & 0.723 & \textbf{0.620} & 0.723 & 0.575 & \textbf{0.820} & 0.554 & \textbf{0.657} \\ \hline
\textbf{GradCon} & \textbf{0.760} & \textbf{0.598} & 0.648 & 0.586 & 0.733 & 0.603 & 0.684 & 0.567 & \textbf{0.784} & \textbf{0.678} & \textbf{0.664}\\ \bottomrule
\end{tabular}\caption{Anomaly detection AUROC results on CIFAR-10.}\label{tab:sota_cifar}\vspace{-0.6cm}
\end{table*} 
\begin{table*}[t]
\scriptsize
\centering
\begin{tabular}{cccccccccccc}
\toprule
 & 0 & 1 & 2 & 3 & 4 & 5 & 6 & 7 & 8 & 9 & Average \\ \hline
OCSVM~\cite{OCSVM} & 0.988 & \textbf{0.999} & 0.902 & 0.950 & 0.955 & 0.968 & 0.978 & 0.965 & 0.853 & 0.955 & 0.951 \\ \hline
KDE~\cite{KDE} & 0.885 & 0.996 & 0.710 & 0.693 & 0.844 & 0.776 & 0.861 & 0.884 & 0.669 & 0.825 & 0.814 \\ \hline
DAE~\cite{DAE} & 0.894 & \textbf{0.999} & 0.792 & 0.851 & 0.888 & 0.819 & 0.944 & 0.922 & 0.740 & 0.917 & 0.877 \\ \hline
VAE~\cite{VAE} & \textbf{0.997} & \textbf{0.999} & 0.936 & 0.959 & \textbf{0.973} & 0.964 & \textbf{0.993} & 0.976 & 0.923 & \textbf{0.976} & 0.970 \\ \hline
PixelCNN~\cite{PixelCNN} & 0.531 & 0.995 & 0.476 & 0.517 & 0.739 & 0.542 & 0.592 & 0.789 & 0.340 & 0.662 & 0.618 \\ \hline
LSA~\cite{abati2019latent} & 0.993 & \textbf{0.999} & \textbf{0.959} & \textbf{0.966} & 0.956 & 0.964 & \textbf{0.994} & \textbf{0.980} & \textbf{0.953} & \textbf{0.981} & \textbf{0.975} \\ \hline
AnoGAN~\cite{GAN} & 0.966 & 0.992 & 0.850 & 0.887 & 0.894 & 0.883 & 0.947 & 0.935 & 0.849 & 0.924 & 0.913 \\ \hline
DSVDD~\cite{DSVDD} & 0.980 & \textbf{0.997} & 0.917 & 0.919 & 0.949 & 0.885 & 0.983 & 0.946 & \textbf{0.939} & 0.965 & 0.948 \\ \hline
OCGAN~\cite{perera2019ocgan} & \textbf{0.998} & \textbf{0.999} & 0.942 & 0.963 & \textbf{0.975} & \textbf{0.980} & 0.991 & \textbf{0.981} & \textbf{0.939} & \textbf{0.981} & \textbf{0.975} \\ \hline
\textbf{GradCon} & 0.995 & \textbf{0.999} & \textbf{0.952} & \textbf{0.973} & 0.969 & \textbf{0.977} & \textbf{0.994} & 0.979 & 0.919 & 0.973 & \textbf{0.973} \\ 
\bottomrule
\end{tabular}\caption{Anomaly detection AUROC results on MNIST.}\label{tab:sota_mnist}\vspace{-0.9cm}
\end{table*}
\subsection{Comparison With State-of-The-Art Algorithms}~\label{subsec:sota}\vspace{-0.1cm}
We evaluate the performance of GradCon which uses the combination of the reconstruction error and the gradient loss as an anomaly score. We compare GradCon with other benchmarking and state-of-the-art algorithms. The AUROC results on CIFAR-10 and MNIST are reported in Table~\ref{tab:sota_cifar} and Table~\ref{tab:sota_mnist}, respectively. Top two AUROC scores for each class are highlighted in bold. GradCon achieves the best average AUROC performance in CIFAR-10 while achieving the second best performance in MNIST by the gap of $0.002$. In Fig. \textcolor{red}{5}, we visualize the histogram of the reconstruction error, the latent loss, and the gradient loss for inliers and outliers to further analyze the state-of-the-art performance of the proposed method. We calculate each loss for all the inliers and the outliers in MNIST. Also, we provide the percentage of overlap calculated by dividing the number of samples in the overlapped region of the histograms by the total number of samples. Ideally, measured errors on each representation should separate the histograms of inliers and outliers as much as possible for effective anomaly detection. The gradient loss achieves the least number of samples overlapped which explains the state-of-the art performance achieved by GradCon. We also evaluate the performance of GradCon in comparison with another state-of-the-art algorithm denoted as GPND~\cite{pidhorskyi2018generative} in fMNIST. In this fMNIST experiment, we change the ratio of outliers in the test set from $10\%$ to $50\%$ and evaluate the performance in terms of AUROC and F1 score. We report the results from the gradient loss (Grad) and GradCon in Table~\ref{tab:sota_fmnist}. GradCon outperforms GPND in all outlier ratios in terms of AUROC. Except for the $10\%$ of outlier ratio, GradCon achieves higher F1 scores than GPND. The results of the gradient loss and GradCon show that the combination of the gradient loss and the reconstruction error improves the performance for all the outlier ratios in terms of AUROC and F1 score. 

\noindent\textbf{Computational efficiency of GradCon} GradCon requires significantly less computational resources compared to other state-of-the-art algorithms. To show the computational efficiency of GradCon, we measure the average inference time per image using a machine with two GTX Titan X GPUs and compare computation time. While the average inference time per image for GPND on fMNIST is $5.72$ \textit{ms}, GradCon takes only $3.08$ \textit{ms} which is around $1.9$ time faster. Also, we compare the number of model parameters for GradCon with that for the state-of-the-art algorithms in Table~\ref{tab:params}. AnoGAN, GPND, and LSA are based on a GAN~\cite{goodfellow2014generative}, an AAD~\cite{makhzani2015adversarial}, and an autoregressive model~\cite{luc2017predicting}, respectively but GradCon is solely based on a CAE. Hence, the number of model parameters for GradCon is approximately $27, 29, 59$ times less than that for AnoGAN, GPND, and LSA, respectively. Most of the state-of-the-art algorithms require additional training of adversarial networks or probabilistic modeling on top of the activation-based representations from the encoder and the decoder. Since GradCon is only based on the reconstruction error and the gradient loss of the CAE, it is computationally efficient even while achieving the state-of-the-art performance.\vspace{-0.25cm}
\begin{table}[t]
\centering
\begin{minipage}{0.59\linewidth}
    \centering
    \scriptsize
    \begin{tabular}{ccccccc}
    \toprule
    \multicolumn{2}{c}{\% of outlier} & 10 & 20 & 30 & 40 & 50 \\ \hline
    \multirow{3}{*}{F1} & GPND & \textbf{0.968} & \textbf{0.945} & 0.917 & 0.891 & 0.864 \\  \cline{2-7} 
    & \textbf{Grad} & 0.964 & 0.939 & 0.917 & 0.899 & 0.870 \\ \cline{2-7}
    & \textbf{GradCon} & 0.967 & \textbf{0.945} & \textbf{0.924} & \textbf{0.905} & \textbf{0.871} \\ \hline
    \multirow{3}{*}{AUC} & GPND & 0.928 & 0.932 & 0.933 & 0.933 & 0.933 \\ \cline{2-7} 
    & \textbf{Grad} & 0.931 & 0.925 & 0.926 & 0.928 & 0.926 \\ \cline{2-7}
    & \textbf{GradCon} & \textbf{0.938} & \textbf{0.933} & \textbf{0.935} & \textbf{0.936} & \textbf{0.934} \\ \bottomrule
    \end{tabular}\caption{Anomaly detection results on fMNIST.}\label{tab:sota_fmnist}
\end{minipage}\hspace{0.3cm}
\begin{minipage}{0.33\linewidth}
    \centering
    \scriptsize
    \begin{tabular}{cc}
    \toprule
    Method & \# of parameters \\ \hline
    AnoGAN & 6,338,176 \\ \hline
    GPND & 6,766,243 \\ \hline
    LSA & 13,690,160 \\ \hline
    \textbf{GradCon} & \textbf{230,721} \\ \bottomrule
    \end{tabular}\caption{Number of model parameters.}\label{tab:params}
\end{minipage}\vspace{-1cm}
\end{table}

\section{Conclusion}\label{sec:conclusion}
\vspace{-0.2cm}We propose using a gradient-based representation for anomaly detection by characterizing model behavior on anomalies. We introduce the geometric interpretation of gradients and derive an anomaly score based on the deviation of gradients from the directional constraint. From thorough baseline analysis, we show the effectiveness of gradient-based representations for anomaly detection in comparison with the activation-based representations. Also, the proposed anomaly detection algorithm, GradCon, which is the combination of the reconstruction error and the gradient loss achieves the state-of-the-art performance in benchmarking image recognition datasets. In terms of the computational efficiency, GradCon has significantly less number of model parameters and shows faster inference time compared to other state-of-the-art anomaly detection algorithms. Given that most of anomaly detection algorithms adopt adversarial training frameworks or probabilistic modelings on activation-based representations, using more sophisticated training frameworks on gradient-based representations remains for future work. 

\clearpage

\bibliographystyle{splncs04}
\bibliography{references}

\clearpage
\appendix
\section{Appendix}
In Section~\ref{suppsec:fmnist}, we compare the performance of GradCon with other benchmarking and state-of-the-art algorithms on fMNIST. In Section~\ref{suppsec:hist}, we perform statistical analysis and highlight the separation between inliers and outliers achieved by using the gradient-based representations in CIFAR-10. In Section~\ref{suppsec:param}, we analyze different parameter settings for GradCon. Finally, we provide additional details on CURE-TSR dataset in Section~\ref{suppsec:cure}.

\subsection{Additional Results on fMNIST}\label{suppsec:fmnist}
We compared the performance of GradCon with other benchmarking and state-of-the-art algorithms using CIFAR-10 and MNIST in Table~\ref{tab:sota_cifar} and~\ref{tab:sota_mnist}.
In Table~\ref{tab:sota_fmnist} of the paper, we mainly focused on rigorous comparison between GradCon and GPND which shows the second best performacne in terms of the average AUROC on fMNIST. In this section, we report the average AUROC performance of GradCon in comparison with that of additional benchmarking and state-of-the-art algorithms using fMNIST in Table~\ref{tab:sota_fmnist_supp}. The same experimental setup for fMNIST described in Section~\ref{subsec:setup} is utilized and the test set contains the same number of inliers and outliers. GradCon outperforms all the compared algorithms including GPND. Given that ALOCC, OCGAN, and GPND are all based on adversarial training to further constrain the activation-based representations, GradCon achieves the best performance in fMNIST only based on a CAE and requires significantly less computations.

\begin{table}[h]
\centering
\scriptsize
\begin{tabular}{ccccccc}
\toprule
Method & ALOCC DR~\cite{sabokrou2018adversarially} & ALOCC D~\cite{sabokrou2018adversarially} & DCAE~\cite{sakurada2014anomaly}  & OCGAN~\cite{perera2019ocgan} & GPND~\cite{pidhorskyi2018generative} & \textbf{GradCon} \\ \hline
AUROC & 0.753    & 0.601   & 0.908 & 0.924 & 0.933 & \textbf{0.934}\\ \bottomrule
\end{tabular}\caption{Average AUROC result of GradCon compared with benchmarking and state-of-the-art anomaly detection algorithms on fMNIST.}\label{tab:sota_fmnist_supp}\vspace{-1.3cm}
\end{table}

\begin{figure}[h]
    \centering
    \includegraphics[width=0.8\linewidth]{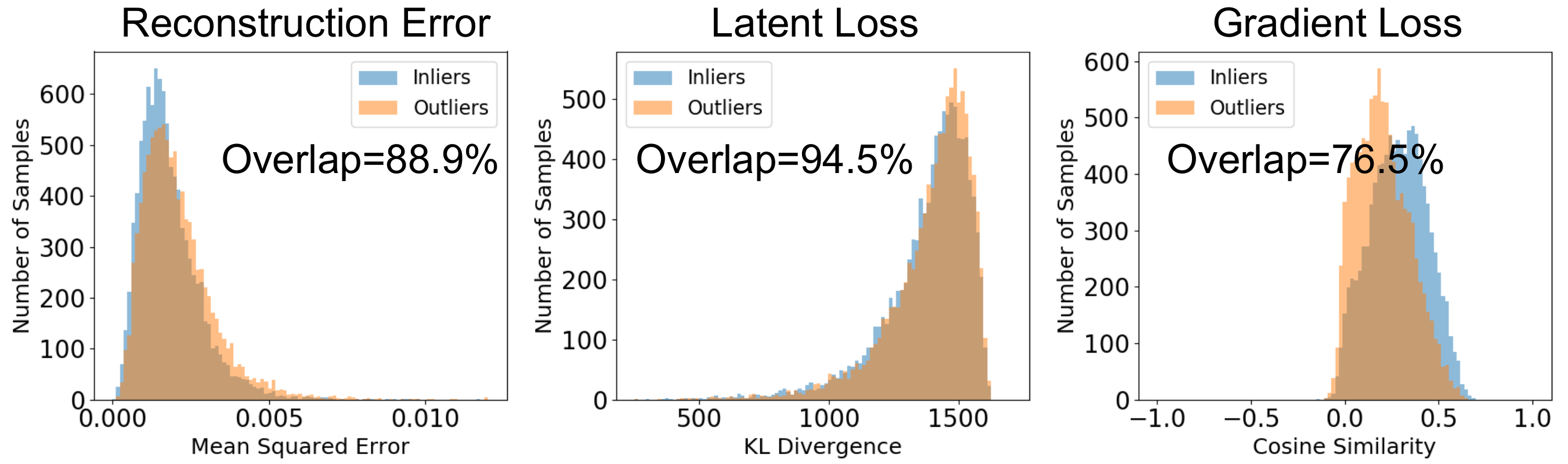}\caption*{Figure 6. Histogram analysis on activation losses and gradient loss in CIFAR-10. For each class, we calculate the activation losses and the gradient loss from inliers and outliers. The losses from all 10 classes are visualized using histograms. The percentage of overlap is calculated by dividing the number of samples in the overlapped region of the histograms by the total number of samples.}\vspace{-1cm}
\end{figure}
\subsection{Histogram Analysis on CIFAR-10}\label{suppsec:hist}
We presented histogram analysis using gray scale digit images in MNIST to explain the state-of-the-art performance achieved by GradCon in Fig.~\textcolor{red}{5}. In this section, we perform the same histogram analysis using color images of general objects in CIFAR-10 to further highlight the separation between inliers and outliers achieved by the gradient-based representations. We obtain histograms for CIFAR-10 through the same procedures that are used to generate histograms for MNIST visualized in Fig.~\textcolor{red}{5}. In Fig.~\textcolor{red}{6}, we visualize the histograms of the reconstruction error, the latent loss, and the gradient loss in CIFAR-10. Also, we provide the percentage of overlap between histograms from inliers and outliers. The measured error on each representation is expected to differentiate inliers from outliers and achieve as small as possible overlap between histograms. The gradient loss shows the smallest overlap compared to other two losses defined in activation-based representations. This statistical analysis also supports the superior performance of GradCon compared to other reconstruction error or latent loss-based algorithms reported in Table~\ref{tab:sota_cifar}.

Comparison between histograms from MNIST visualized in Fig.~\textcolor{red}{5} and those from CIFAR-10 shows that the gradient loss is more effective when data becomes complicated and challenging for anomaly detection. In MNIST, simple low-level features such as curved edges or straight edges can be class discriminant features for anomaly detection. On the other hand, CIFAR-10 contains images with richer structure and features than MNIST. Therefore, normal and abnormal data are not easily separable and the overlap between histograms is significantly larger in CIFAR-10 than MNIST. In CIFAR-10, the overlap of the gradient loss is smaller than the second smallest overlap of the reconstruction error by $12.4\%$. In MNIST, the overlap of the gradient loss is smaller than the second smallest overlap by $5.7\%$. GradCon also outperforms other state-of-the-art methods by a larger margin of AUROC in CIFAR-10 compared to MNIST. The overlap and performance differences show that the contribution of the gradient loss becomes more significant when data is complicated and challenging for anomaly detection. 
\begin{figure}[t]
    \centering
    \includegraphics[width=0.5\linewidth]{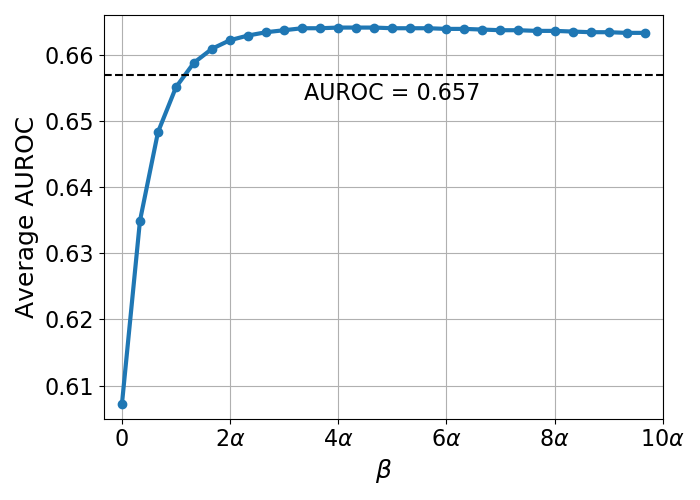}\vspace{-0.2cm}\caption*{Figure 7. Average AUROC results with different $\beta$ parameters in CIFAR-10. $\alpha = 0.03$ is utilized to train the CAE. The dotted line (average AUROC = 0.657) indicates the performance of OCGAN which achieves the second best performance in CIFAR-10.}\vspace{-0.3cm}
\end{figure}

\subsection{Parameter Setting for the Gradient Loss}\label{suppsec:param}
We analyze the impact of different parameter settings on the performance of GradCon. The final anomaly score of GradCon is given as $\mathcal{L} + \beta \mathcal{L}_{grad}$, where $\mathcal{L}$ is the reconstruction error and $\mathcal{L}_{grad}$ is the gradient loss. While we use $\alpha$ parameter to weight the gradient loss and constrain the gradients during training, we observe that the gradient loss generally shows better performance as an anomaly score than the reconstruction error. Hence, we use $\beta = n \alpha$, where n is constant, to weight the gradient loss more for the anomaly score. We evaluate the average AUROC performance of GradCon with different $\beta$ parameters using CIFAR-10 in Fig.~\textcolor{red}{7}. In particular, we change the scaling constant, $n$, to change $\beta$ in the $x$-axis of the plot. The performance of GradCon improves as we increase $\beta$ in the range of $\beta = \left[0, 2\alpha\right]$. Also, GradCon consistently achieves state-of-the-art performance across a wide range of $\beta$ parameter settings when $\beta \geq 1.67 \alpha$. To be specific, GradCon always outperforms OCGAN which achieves the second best average AUROC performance of $0.657$ in CIFAR-10 when $\beta \geq 1.67\alpha$. This analysis shows that GradCon achieves the best performance in CIFAR-10 across a wide range of $\beta$. 

\subsection{Additional Details on CURE-TSR Dataset}\label{suppsec:cure}
We visualize traffic sign images with 8 different challenge types and 5 different levels in Fig.~\textcolor{red}{8}. Level 5 images contain the most severe challenge effect and level 1 images are least affected by the challenging conditions. Since level 1 images are perceptually most similar to the challenge-free image, it is more challenging for anomaly detection algorithms to classify level 1 images as outliers. The gradient loss from CAE + Grad outperforms the reconstruction error from CAE in all level 1 challenge types. This result shows that the gradient loss consistently outperforms the reconstruction error even when inliers and outliers become relatively similar under mild challenging conditions. 

\begin{figure}
    \hspace{1.7cm}
    \includegraphics[width=0.6\linewidth]{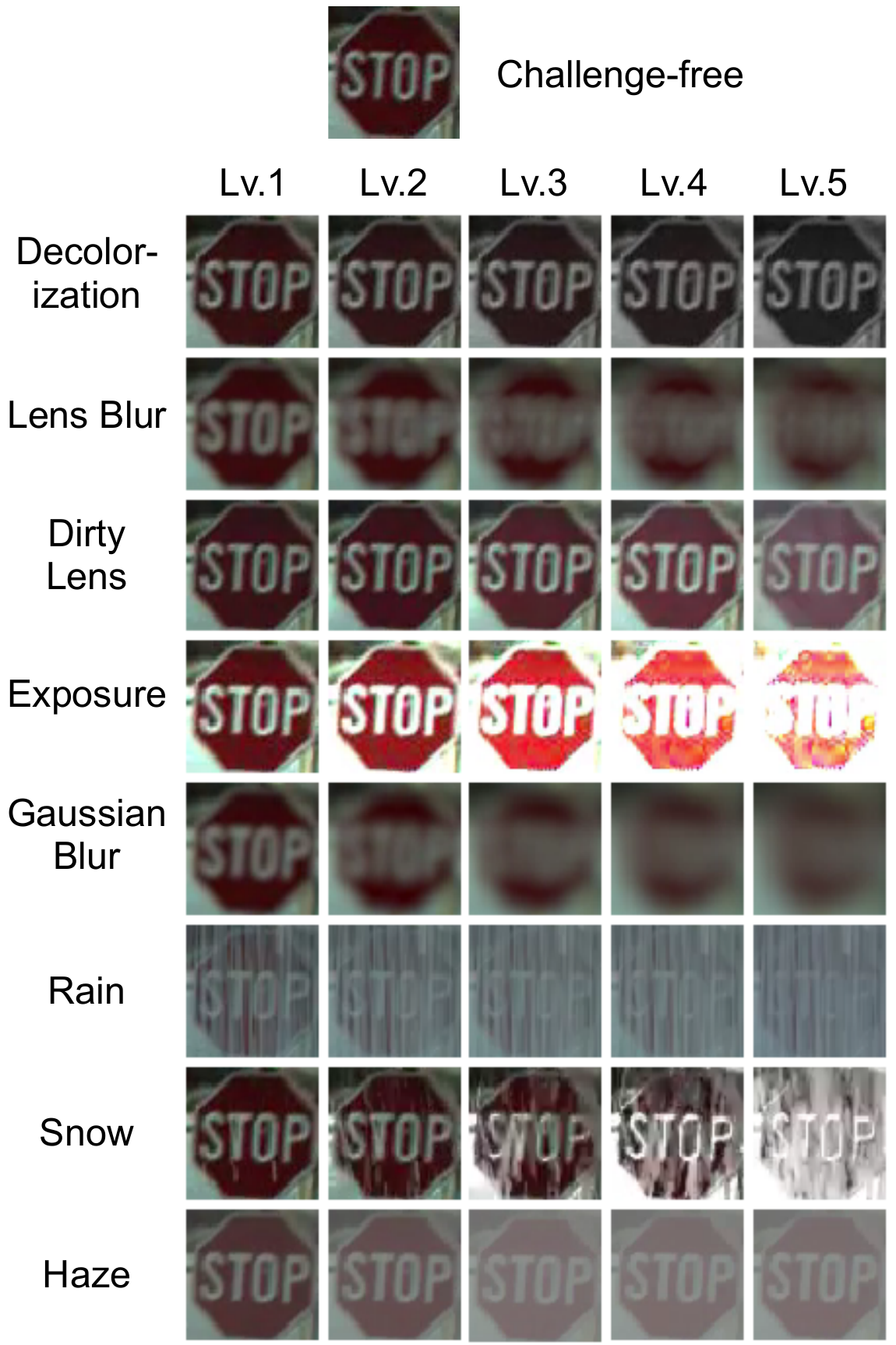}\caption*{Figure 8. A challenge-free stop sign and stop signs with 8 different challenge types and 5 different challenge levels. Challenging conditions become more severe as the level becomes higher.}
\end{figure}

\end{document}